# Expert and Non-Expert Opinion about Technological Unemployment

**Toby Walsh**
UNSW Sydney | Data61 | TU Berlin

**Abstract**

There is significant concern that technological advances, especially in Robotics and Artificial Intelligence (AI), could lead to high levels of unemployment in the coming decades. Studies have estimated that around half of all current jobs are at risk of automation. To look into this issue in more depth, we surveyed experts in Robotics and AI about the risk, and compared their views with those of non-experts. Whilst the experts predicted a significant number of occupations were at risk of automation in the next two decades, they were more cautious than people outside the field in predicting occupations at risk. Their predictions were consistent with their estimates for when computers might be expected to reach human level performance across a wide range of skills. These estimates were typically decades later than those of the non-experts. Technological barriers may therefore provide society with more time to prepare for an automated future than the public fear. In addition, public expectations may need to be dampened about the speed of progress to be expected in Robotics and AI.

## 1  Introduction

The World Economic Forum has predicted that we are the beginning of a Fourth Industrial Revolution in which developments in areas like Robotics and Artificial Intelligence will transform the nature of our economies and eliminate many current occupations [WEF 2014]. At the same time, these technologies will also create many new occupations. It remains an interesting question whether more or fewer jobs will be created than destroyed. In the past, more jobs have been created than destroyed but this may be the case in the future as we are likely to have fewer and fewer advantages over the machines. Whatever the case, it is likely that the new occupations created will require different skills to those destroyed. For instance, autonomous vehicles will probably be commonplace on our roads within the next few decades. Taxi and truck drivers will therefore need other skills than just the ability to drive if they are to remain employed. It is thus an important question for our societies in preparing for this future to understand the occupations at risk of automation.

## 2  Background

In 2013, a study by Frey and Osborne estimated that 47% of total employment in the United States was under risk of automation in the next two decades [Frey and Osborne 2013]. Ironically, the study used Machine Learning to predict occupations at risk. Even the occupation of predicting occupations at risk from automation has been partially automated. Subsequent studies have reached similar conclusions. For instance, similar analysis has estimated that 40% of total employment in Australia is at risk of automation [Durrant-Whyte *et al.* 2015], and even larger figures for developing countries like China at 77% and India at 69% [Frey *et al.* 2016]. Frey and Osborne suggested three barriers to automation: occupations requiring complex perception or manipulation skills, occupations requiring creativity, and occupations requiring social intelligence. Computers are significantly challenged in these three areas at present and may remain so for some time to come.

Frey and Osborne's study used a training set of 70 occupations from the O*Net database of U.S. occupations. This training set was hand labelled by a small group of economists and Machine Learning researchers at a workshop held in the Oxford University Engineering Sciences Department. Classification was binary. Each occupation was classified either at risk in the next two decades from automation or not. Labels were only assigned to occupations where there was confidence in the classification.

We do not wish to discuss here whether the O*Net database provides features adequate to extrapolate to the full set of 702 occupations. This is a difficult question to address as we do not have a gold standard of occupations *actually* at risk. Their classifier did, however, perform well on the training set with a precision (positive predictive value) for occupations at risk of automation of 94%, a sensitivity of 81%, and a specificity of 94%.

We focus instead on the training set of 70 occupations used in [Frey and Osborne 2013]. This study hand labelled 37 of these 70 occupations as being at risk of automation (53%). The final accuracy of the classification of 702 occupations depends critically on the accuracy with which this smaller training set was hand labelled. This training set was chosen as it could be classified "with confidence". We therefore gave this training set to three much larger groups to classify: experts in AI, experts in Robotics and, as a comparison, non-experts interested in the future of AI. In total over, we sampled over 300 experts and 500 non-experts. Our survey is the largest of its kind every performed.

## 3 High level machine intelligence

In addition to classifying the training set, we asked both the experts and the non-experts to estimate when computers might be expected to achieve a high–level of machine intelligence (HLMI). This was defined to be when a computer might be able to carry out most human professions at least as well as a typical human. In 2012/2013, Vincent C. Müller and Nick Bostrom surveyed 170 people working in AI to predict when HLMI might be achieved [Müller and Bostrom 2014]. As there is significant uncertainty as to when HLMI might be achieved, they asked when the probability of HLMI would be 10%, 50% and 90%. The median response for a 10% probability of HLMI was 2022, for a 50% probability was 2040, and for a 90% probability was 2075. We wanted to see if people who were more cautious at predicting when HLMI was likely to be achieved were also more cautious at predicting occupations at risk of automation.

We also wished to update and enlarge upon Müller and Bostrom's survey. Given some of the high profile advances made recently in subareas of AI like Deep Learning [LeCun *et al.* 2015]*,* it might be expected that HLMI would be predicted sooner now than back in 2012/2013. We also wanted to survey a much larger sample of experts in AI and Robotics than Müller and Bostrom.

Only 29 of the 170 who answered Müller and Bostrom's survey were leading experts in AI, specifically 29 members of the 100 must cited authors in AI as ranked by Microsoft Academic Research. The largest group in their survey were 72 participants of a conference in Artificial General Intelligence (AGI). This is a specialized area in AI where researchers are focused on the question of building general intelligence. Much research in AI is, by comparison, focused on programming computers to do very specialized tasks like playing Go [Silver *et al.* 2016] or interpreting mammograms [Patel *et al.* 2017] and not on building general purpose intelligence.

Researchers in AGI might be expected to be predisposed to the early arrival of HLMI. Indeed the AGI group were the most enthusiastic to complete Müller and Bostrom's survey. 64% of the delegates from this AGI conference completed the survey, compared to an overall response rate of just 31%. In addition, the AGI group typically pre-dicted HMLI would arrive earlier than the other respondents to the survey. We conjectured that experts in AI and Robotics not focused on AGI would be more cautious in their predictions.

More recently in March 2016, Oren Etzioni wanted to test a similar hypothesis about Müller and Bostrom's results [Etzioni 2016]. To do so, he sent out a survey to 193 Fellows of the Association for the Advancement of Artificial Intelligence (AAAI). In total, 80 Fellows responded (41% response rate). Respondents included many leading researchers in the field like Geoff Hinton, Ed Feigenbaum, Rodney Brooks, and Peter Norvig.

Unfortunately, Etzioni's survey asked a different and simpler question ("When do you think we will achieve Superintelligence?" where Suprintelligence is defined to be "an intellect that is much smarter than the best human brains in practically every field, including scientific creativity, general wisdom and social skills"). Etzioni's survey also only offered 4 answers to the question of when Superintelligence would be achieved (in next 10 years, 10-25 years, more than 25 years, never).

It is difficult to compare the results of Etzioni's survey with Müller and Bostrom's. None of the AAAI Fellows responding selected "in the next 10 years", 7.5% selected "in the next 10-25 years", 67.5% selected "in more than 25 years", and the remaining 25% selected "never". If Etzioni's question is equated with Müller and Bostrom's question about a 90% probability of HLMI then the responses of the two surveys appear to be similar. However, it is very difficult to draw many conclusions given the rather ambiguous question, and the larger granularity on the answers.

## 4 Methods

Our survey was performed between 20[th] January and 5[th] February 2017. The survey involved three distinct groups. The first group were authors from two leading AI conferences: the annual Conference of the Association for the Advancement of Artificial Intelligence (AAAI-2015), and the International Joint Conference on Artificial Intelligence (IJCAI-2011). Both conferences are highly selective and publish some of the best new work in AI. 200 authors from this group completed our survey.

The second group consisted of IEEE Fellows in the IEEE Robotics & Automation Society and authors of a leading Robotics conference: the IEEE International Conference on Robotics and Automation (ICRA-2016). This is also a highly selective conference that publishes some of the best work in Robotics. We sent out questionnaires to this second group till we had at least 100 replies. In total, 101 people from this group completed the survey.

The third and final group surveyed were readers of an article from the website "The Conversation". This Australian and British website publishes news stories and expert opinion from the university sector, and is partnered with Reuters and the Press Association. The article containing the like to the survey was entitled *"Know when to fold 'em: AI beats world's top poker players"*. The article discussed the recent victory of the CMU Libratus poker program against some top human players. It used this as an introduction to the Frey and Osborne report on tasks that could be automated. It ended by inviting readers to help determine the "wisdom of the crowd" by completing the survey. There were 548 responses in this third group.

The readers of The Conversation have the following geographical distribution: 36% Australia, 29% United States, 7% United Kingdom, 4% Canada, and 24% rest of the world. It is reasonable to suppose that they most are not experts in AI & Robotics, and that they are unlikely to be publishing in the top venues in AI and Robotics like IJCAI, AAAI or ICRA. They are educated (85% have an undergraduate degree or higher), young (more than a third are 34 or under, 59% are under 44 and just 11% are 65 or older), mostly employed or in higher education (more than two thirds are employed and one quarter are in or about to enter higher education) and relatively affluent (40% reported an annual income of $100,000 or more).

The questionnaire itself had 8 questions. The first 7 questions asked respondents to classify 10 occupations from the training set, whilst the last asked for estimates when HLMI might arrive. The first question asked for a classification of the 5 occupations most at risk from automation according to Frey and Osborne's classifier as well as the 5 occupations least likely to be at risk. To help respondents, a link was provided next to each occupation describing the work involved and the skills required. The second question in our survey asked for a classification of the next 5 occupations most at risk from automation according to Frey and Osborne's classifier and the next 5 occupations least likely, and so on till the seventh and penultimate question. Within each of the 7 questions, the 10 occupations were presented in a random order. Our intent was to make the initial questions as easy as possible to answer. In this way, we hoped that participants would not give up early, and might be better prepared for the potentially more difficult classifications later in the survey.

The 8th and final question asked for an estimate of when there was a 10%, 50% and 90% chance of HLMI. The options presented were: 2025, 2030, 2040, 2050, 2075, 2100, after 2100, and never. To compute the median response, we interpolated the cumulative distribution function between the two nearest dates.

## 5 Results

The results are summarized in Table 1. The experts in Robotics were most cautious, predicting a mean and median of 29.0 out of the 70 occupations in the training set at risk from automation (95% confidence interval of 27.0 to 31.0 occupations at risk). The experts in AI were slightly less cautious predicting a mean of 31.1 occupations at risk and a median of 33 (95% confidence interval of 29.6 to 32.6 occupations at risk).

The difference in means between the Robotics and AI experts does not appear to be statistically significant. A two-sided student t-test on the number of occupations predicted at risk of automation failed to reject the null hypothesis that the population means were equal at the 95% level ($p$ value of 0.096).

| Group | Sample size ($n$) | Predicted Number of Occupations Likely at Risk of Automation (out of 70) | | | |
|---|---|---|---|---|---|
| | | Mean | Median | Standard deviation | Confidence interval |
| Robotics experts | 101 | 29.0 | 29 | 10.1 | (27.0, 31.0) |
| AI experts | 200 | 31.1 | 33 | 10.8 | (29.9, 32.6) |
| Non-experts | 473 | **36.5** | **37** | 10.9 | (35.6, 37.5) |

**Table 1. Descriptive statistics about number of occupations predicted to be at risk of automation in next two decades.** Confidence intervals are at the 95% level.

The non-experts in our survey typically predicted significantly more occupations were at risk of automation than the experts. They predicted a mean of 36.5 occupations at risk of automation and a median of 37 (the 95% confidence interval is from 35.6 to 37.5 occupations at risk).

The differences between the predictions by the non-experts of the number of occupations at risk of automation and those of either the Robotics or the AI experts appear to be extremely significant statistically. Two-sided student t-tests rejected the null hypothesis that the population means for the non-experts and the experts in Robotics were equal, and the null hypothesis that the population means for the non-experts and the experts in AI were equal (both $p$ values less than 0.0001).

The prediction by the non-experts in our survey of the number of occupations at risk of automation of a median of 37 occupations at risk is identical to the 37 occupations labelled at risk in the original training set in the original Frey and Osborne study.

At the end of the survey, we asked participants to estimate when there was a 10%, 50% and 90% probability of HLMI. This repeats a question asked in the original Müller and Bostrom survey. Also, as in Müller and Bostrom's survey, we defined HLMI to be when a computer can carry out most human professions at least as well as a typical human.

The results of this question are summarized in Figure 1. The Robotics and AI experts typically predicted that HLMI was several decades further away than the non-experts. Again, there was little to distinguish between the AI and Robotics experts themselves, but they were much more cautious than the non-experts in their predictions. The experts typically predicted HLMI was several decades further away than the non-experts.

For a 90% probability of HLMI, the median prediction of the experts in Robotics was 2118, and 2109 for the experts in AI. By comparison, the median prediction of the non-experts for a 90% probability of HLMI was just 2060, around half a century earlier. For a 50% probability of HLMI, the median prediction of the Robotics experts was 2065, and 2061 for the AI experts. This compares with the non-experts whose median prediction for a 50% probability of HLMI was 2039, over two decades earlier. Finally, for a 10% probability of HLMI, the median prediction of the Robotics experts was 2033, and 2035 for the AI experts. By comparison, the median prediction of the non-experts for a 10% probability of HLMI was 2026, nearly a decade earlier.

The predictions for the number of occupations under risk of automation were consistent with the predictions of when HLMI might be achieved. See the clear trend in Figure 1/d. Respondents who predicted a later date for HLMI typically predicted fewer occupations at risk of automation. Similarly respondents who predicted an earlier date for HLMI typically predicted more occupations at risk of automation. The AI and Robotics experts typically predicted later dates for HLMI and fewer occupations at risk. On the other hand, the non-experts typically predicted earlier dates for HLMI and more occupations at risk of automation.

The respondents in Müller and Bostrom's study were closest in their predictions of when HLMI might be achieved to the group of non-experts in our survey. For a 10% probability of HLMI, Müller and Bostrom's study had a median prediction of 2022, and 2040 for a 50% probability of HLMI. For a 10% probability of HLMI, the non-experts in our study had a median prediction of 2026, and of 2039 for a 50% probability of HLMI. However, for a 90% probability of HLMI, our non-experts were more optimistic than the respondents in Müller and Bostrom's study. The median prediction for a 90% probability for HLMI by the non-experts in our survey was 2060, compared to a median of 2075 in Müller and Bostrom's study.

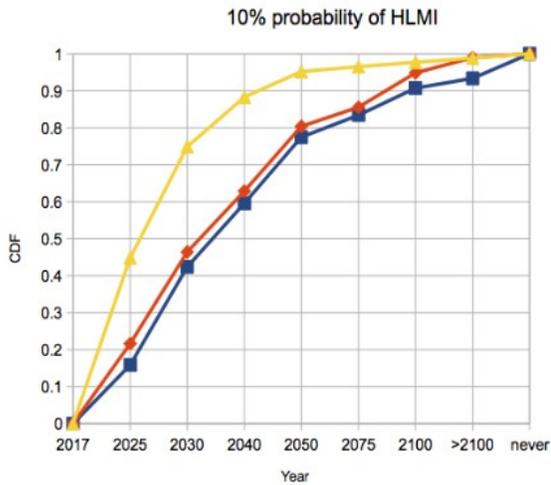
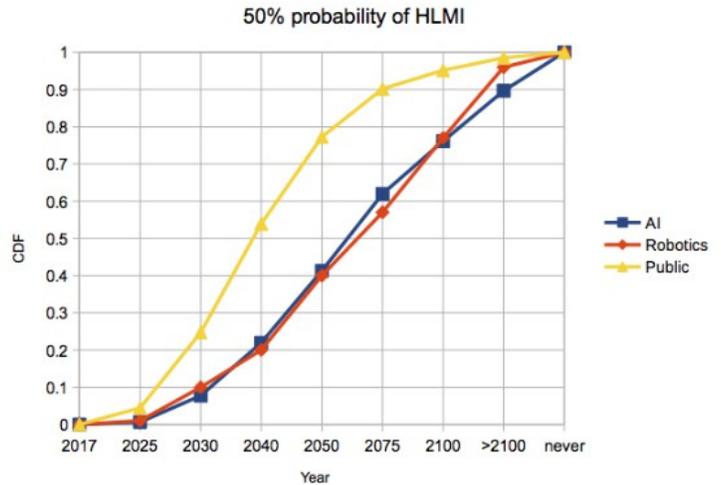

(a) (b)

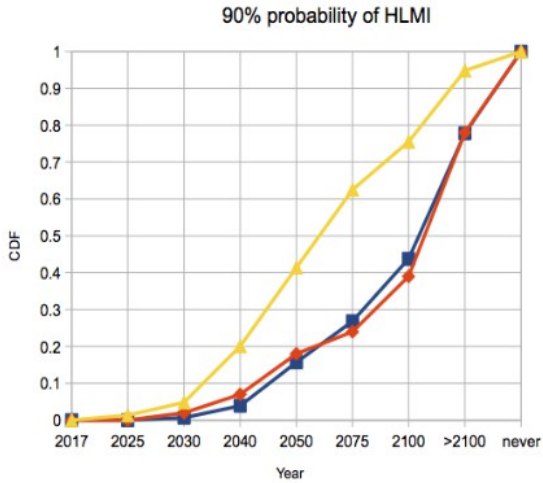
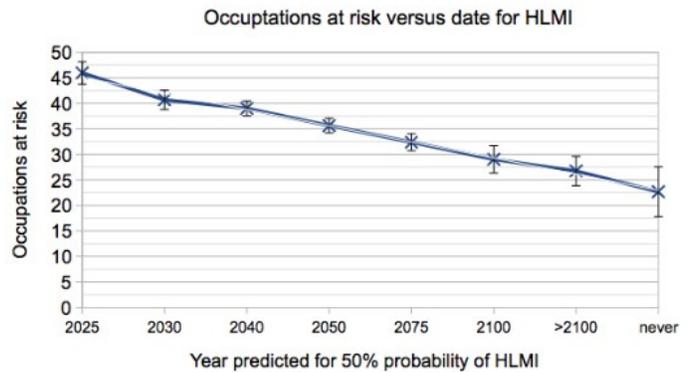

(c) (d)

**Fig. 1. Predictions of Date of High Level Machine Intelligence (HLMI).** This was defined to be when a computer might be able to carry out most human professions at least as well as a typical human. **(a)** Cummulative Distribution Function (CDF) for a 10% probability of HLMI, **(b)** Cummulative Distribution Function for a 50% probability of HLMI, **(c)** Cummulative Distribution Function for a 90% probability of HLMI. **(d)** Mean number of occupations predicted at risk of automation against year predicted for a 50% probability of HLMI. Error bars give 95% confidence interval.

# 6   Discussion

Our results suggest that experts in Robotics and AI are more cautious than non-experts in their prediction of the number of occupations at risk of automation in the next decade or two. The experts were also more cautious than the training set used in Frey and Osborne's study. This caution can be explained by their expectation that HLMI may take several decades longer than the public expects. We did not find any significant differences between the predictions of the experts in Robotics and the experts in AI. Despite being more cautious, both groups of experts still predicted a large fraction of occupations were at risk of automation in the next couple of decades.

There are many other factors that need to be taken into account in deciding the impact that automation might have on employment: we must also take account of the economic growth fueled by productivity gains, the new occupations created by technology, the effects of globalization, changes in demographics and retirement, and much else. It remains an important open question if there will be an overall net gain or loss of jobs as a result. This is a matter that society must seriously consider further. There are many actions possible to reduce the negative impacts of automation. We should, for instance, look to augment rather than replace humans in roles where this is possible.

Even in occupations where humans look set to be displaced, our survey holds out some hope. Whilst the potential disruptions may be large, there could be more time to adapt to them than the public fear. Our study also suggests that more effort needs to be invested in managing the public's expectation about the rate of progress being made in Robotics and AI, and of the many technical obstacles that must be overcome before some occupations can be automated. Robotics and AI remain challenged in several fundamental areas like manipulation, common sense reasoning and natural language understanding. Funding for AI research has suffered "winters" in the past where public expectations did not match actual progress [Hendler 2008]. We should be careful to avoid this in the future.